\documentclass[10pt,twocolumn,letterpaper]{article}

\usepackage{wacv}
\usepackage{times}
\usepackage{epsfig}
\usepackage{graphicx}
\usepackage{amsmath}
\usepackage{amssymb}
\DeclareMathOperator{\E}{\mathbb{E}}
\def\wacvPaperID{747} % Enter the WACV Paper ID here

\wacvfinalcopy % *** Uncomment this line for the final submission
\ifwacvfinal
\fi

\ifwacvfinal
\usepackage[breaklinks=true,bookmarks=false]{hyperref}
\else
\usepackage[pagebackref=true,breaklinks=true,colorlinks,bookmarks=false]{hyperref}
\fi

\ifwacvfinal
\else
\fi

\begin{document}

%%%%%%%%% TITLE
\title{ StressNet: Detecting Stress in Thermal Videos}

\author{Satish Kumar\textsuperscript{\dag} ~ A S M Iftekhar\textsuperscript{\dag} ~ Michael Goebel\textsuperscript{\dag} ~ Tom Bullock\textsuperscript{\ddag} ~ Mary H. MacLean\textsuperscript{\ddag} ~ Michael B. Miller\textsuperscript{\ddag} \\
~ Tyler Santander\textsuperscript{\ddag} ~ Barry Giesbrecht\textsuperscript{\ddag} ~ Scott T. Grafton\textsuperscript{\ddag} ~ B.S. Manjunath\textsuperscript{\dag}\\
{\small University of California Santa Barbara \textsuperscript{\dag}Dept. of Electrical \& Computer Engineering} \\
{\small \textsuperscript{\ddag}Dept. of Psychological and Brain Sciences and Institute for Collaborative Biotechnologies}\\
{\small\{satishkumar@ece, iftekhar@ece, mgoebel@ece, tombullock@, marymaclean@, mbmiller@,} \\
{\small t.santander@, giesbrecht@, stgrafton@, manj@ece\}.ucsb.edu}
}

\maketitle
%\thispagestyle{empty}

%%%%%%%%% ABSTRACT
%%%%%%%%% ABSTRACT
\begin{abstract}

Precise measurement of physiological signals is critical for the effective monitoring of human vital signs. Recent developments in computer vision have demonstrated that signals such as pulse rate and respiration rate can be extracted from digital video of humans, increasing the possibility of contact-less monitoring. This paper presents a novel approach to obtaining physiological signals and classifying stress states from thermal video. The proposed network--"StressNet"--features a hybrid emission representation model that models the direct emission and absorption of heat by the skin and underlying blood vessels. This results in an information-rich feature representation of the face, which is used by spatio-temporal network for reconstructing the ISTI ( Initial Systolic Time Interval : a measure of change in cardiac sympathetic activity that is considered to be a quantitative index of stress in humans). The reconstructed ISTI signal is fed into a stress-detection model to detect and classify the individual’s stress state (i.e. stress or no stress). A detailed evaluation demonstrates that StressNet achieves estimated the ISTI signal with 95\% accuracy and detect stress with average precision of 0.842. The source code is available on Github\footnotemark[\value{footnote}]

\end{abstract}
\vspace{-0.5cm}

\footnotetext{\url{https://github.com/UCSB-VRL/StressNet-Detecting-stress-from-thermal-videos}}
\paragraph*{Keywords:}
\noindent Stress Detection, rPPG, ISTI signal, physiological signal measurement, ECG, ICG, Deep learning model
%%%%%%%%% BODY TEXT
%INTRODUCTION
\section{Introduction}\label{sec:intro}

As the world has come to a standstill due to a deadly pandemic~\cite{roser2020coronavirus}, the need for non-contact, non-invasive health monitoring systems has become imperative. Remote photoplethysmography (rPPG) provides a way to measure physiological signals remotely without attaching sensors, requiring only video recorded with a high-resolution camera to measure the physiological signals of human health. Much of the recent research in the area of rPPG~\cite{wang2016algorithmic} has focussed on leveraging modern computer vision based systems ~\cite{chen2018deepphys,yu2019remote,mcduff2014remote,bousefsaf2013remote} to monitor human vitals such as heart rate and breathing rate. More recent work has expanded these methods to detecting more complex human physiological signals and using them to classify stress states ~\cite{yu2019remote,mcduff2014remote,bousefsaf2013remote}.

\begin{figure}[t]
\begin{center}
\includegraphics[width=\linewidth]{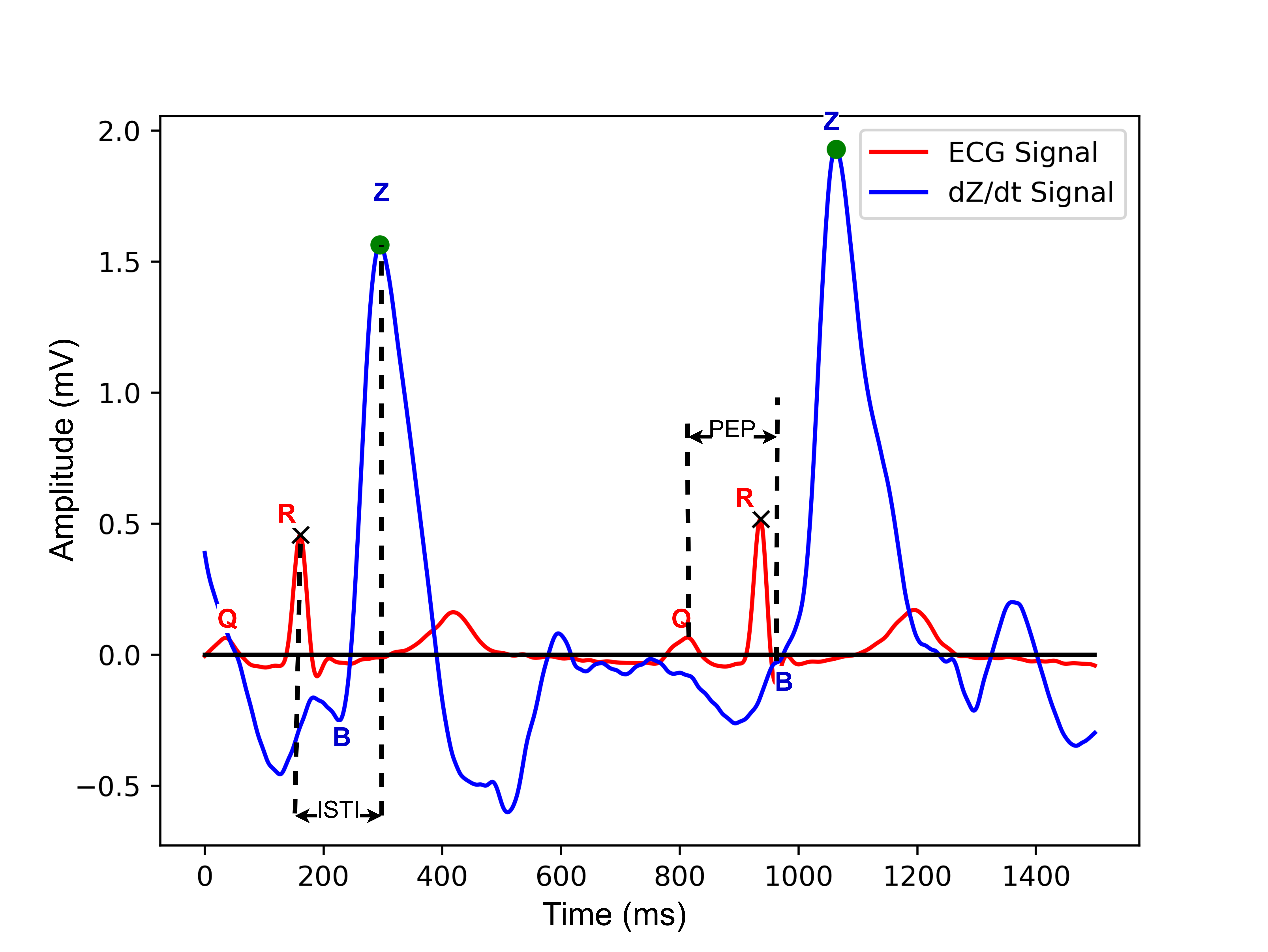}
\end{center}
   \caption{ Example of ECG and $\partial Z / \partial t$ waveforms computed from the present data. $\partial Z / \partial t$ represents the change in impedance recorded by ICG (Z) signal with time. After each ECG peak value there exists an $\partial Z / \partial t$ peak value. The time difference between these two values is known as the initial systolic time interval (ISTI). %\vspace{-0.5cm}
   }
\label{fig:ecg-icg}
\end{figure} 

\begin{figure*}[t]
\begin{center}
\includegraphics[width=\linewidth,height=5.5cm]{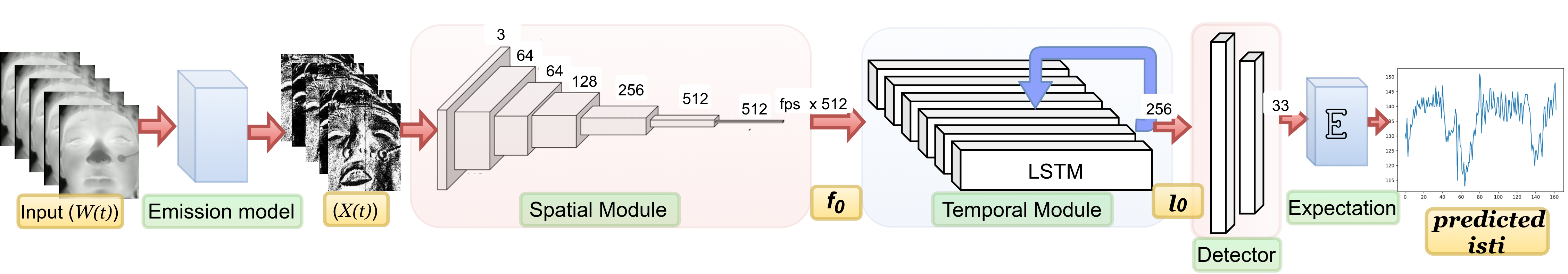}
\end{center}
   \caption{Model Architecture. Green boxes are the different modules of the model. Yellow  boxes are the variables throughout the model. The Emission model processes the raw input data which is then fed into spatial and temporal modules. The Detector network predicts ISTI value for each of the frames from the output of these modules. This ISTI signal is used as input in our stress detection network.
  % \vspace{-0.5cm}
   }
\label{fig:model_architecture}
\end{figure*}

    Whereas all recent datasets for rPPG only collect electrocardiogram (ECG) as the cardiovascular ground truth signal, here we recorded both ECG and impedance cardiography (ICG). ICG is a noninvasive technology measuring total electrical conductivity of the thorax. It is the measure of change in impedance due to blood flow.
    With these two signals, we have the ability to estimate more accurate quantifiers of cardiac sympathetic activity~\cite{silviarz}. Two common metrics are pre-ejection time (PEP) and initial systolic time interval (ISTI).

PEP is the strongest cue for cardiac sympathetic activity. It is defined as the interval from the onset of left ventricular depolarization, reflected by the Q-wave onset in the ECG, to the opening of the aortic valve, reflected by the B-point in the $\partial Z / \partial t$ (derivative of ICG or Z) signal~\cite{riese2003large, willemsen1996ambulatory} as can be seen in Figure~\ref{fig:ecg-icg}. Unfortunately, measuring PEP from ECG and  $\partial Z / \partial t$ signals is quite difficult as the Q and B points that define PEP are subtle and very difficult to pinpoint~\cite{seery2016preejection, van2014comparison}. Accuracy of methods to estimate PEP are low and precision differs highly among studies~\cite{lozano2007b, seery2016preejection}. Instead, ISTI can be used as a reliable index of cardiac sympathetic activity~\cite{silviarz}. ISTI is a straightforward calculation defined as the time difference between the consecutive peaks of ECG and $\partial Z / \partial t$.
    ISTI is considered a strong index of myocardial contractility ~\cite{van2014comparison,meijer2008method} and numerous efforts have shown that ISTI can be used to analyze different physiological phenomena e.g. stress, blood pressure~\cite{meijer2008method, wilde1981evaluation,van2013estimated,meijer2008method}.
    
    Here we introduce StressNet, a non-contact based approach to estimating ISTI.  To the best of our knowledge this approach is the first of it's kind.  StressNet leverages the ISTI signal to classify whether a person is experiencing stress or not. To estimate the ISTI signal, a spatial-temporal deep neural network has been developed along with an emission representation model. Other physiological signals like heart rate (HR) or heart rate variability (HRV) cannot measure the changes in contractility, which are influenced by sympathetic, but not by parasympathetic activity, in humans~\cite{newlin1979pre}.

    Recently a number of studies have applied deep learning methods to the detection of HR or HRV from face videos ~\cite{yu2019remote,chen2018deepphys, hsu2017deep,niu2018synrhythm}. Most of these methods either fail to correctly identify the peak information in ECG or do not properly exploit the temporal relations in the face videos ~\cite{yu2019remote}. Recent work by ~\cite{yu2019remote} has developed a spatial-temporal deep network to measure rPPG signals such as heart rate variability (HRV) and average heart rate (AHR). Although these measurements are important, %we argue that ISTI signals are better indicator on estimating physiological phenomenon like stress.
    we show that in our experimental setup, the ground-truth ISTI signals allow for more accurate classification of stress state than AHR or HRV.
    %Our extensive experimental results have shown that ISTI acts as a strong prior while detecting stress. 
    
    In addition, thermal images mitigate some privacy concerns because the true likeness of the face is not being stored unlike RGB based models~\cite{gunther2017remote}.
    %In addition, most of the deep learning methods such as~\cite{yu2019remote, chen2018deepphys} use RGB videos to estimate physiological signals. Although RGB video recordings are relatively cheap and convenient to collect, they can be problematic to analyze due to changes in ambient light levels which can adversely affect the estimation of ISTI or HRV. Since StressNet deals with thermal images, this eliminates the issue of privacy with certain applications~\cite{gunther2017remote}. To circumvent these issues, here we trained StressNet on a dataset of face videos captured using a thermal camera.   
    
    StressNet is an end-to-end spatial-temporal network that estimates ISTI signal  and attempts to classify stress states based on thermal video recordings of the human face. An extensive analysis of the detailed dataset developed for this work has shown correlation between the estimated signal and ground truth. The effectiveness of this predicted ISTI signal is further validated by the model's ability to accurately classify an individual's stress state.

\paragraph{Technical Contributions:}
\begin{itemize}
  \item An emission representation module is proposed that can be applied to infrared videos to model variations in emitted radiation due to motion of blood and head movements.
  \item A spatial temporal deep neural network is developed to estimate ISTI. 
 \item A simple classifier is then trained to estimate the stress level from the computed ISTI signal. To the best of our knowledge this is the first attempt to directly estimate ISTI and stress from thermal video. 
  %\item With the extensive analysis, we have shown how this estimated ISTI signal can be used to detect physiological phenomenon like stress.
\end{itemize}

\section{Related Works}
%\subsection{Datasets and Tasks}
ISTI has been proposed as an effective, quantitative measure of psychological and physiological stress ~\cite{kelsey2012beta, prasad2017detection,forouzanfar2019automatic,hinnant2011developmental,brenner2011pre}. Measurement of ISTI requires both the ECG and $\partial Z / \partial t$ signals. Heart rate variability has also been used in several studies to index psychological and physiological stress~\cite{yu2019remote,chen2018deepphys, hsu2017deep,niu2018synrhythm}. Different camera modalities have also been used, namely, infrared, visible RGB, and five-channel multi-spectral~\cite{bara2020deep, chen2018deepphys, yu2019remote, bousefsaf2013remote, yu2020autohr}.
A distinction is also made between research that takes place under laboratory and real-world settings. In the latter, environmental variables can complicate the detection and/or estimation task. 

%\subsection{Methodology}

While no other works have included ISTI estimation or the ICG signal in their frameworks, the common video and ECG inputs lend themselves to similar network designs. 

Several works for estimation of heart rate rely solely upon registration and classical signal processing techniques. For example, work from~\cite{li2014remote} registered a region of interest on the face, took the mean of the green channel, and passed that signal through a bandpass filter to estimate the heartbeat signal. At its time of publication in 2014, it achieved state-of-the-art performance on the MAHNOB-HCI dataset with a mean-squared error of 7.62 bpm \cite{li2014remote}.

Several studies have investigated heart rate variability estimation, using a variety of sensor types \cite{mcduff2014remote,bousefsaf2013remote,monkaresi2016automated,blocher2017online}. 

The first end-to-end trainable neural network for rPPG was DeepPhys~\cite{chen2018deepphys}. It replaced the classical face detection methods with a deep learning attention mechanism. Temporal frame differences are fed to the model, in addition to the current frame, to allow the network to learn motion compensation. 

A more recent model built on DeepPhys is PhysNet~\cite{yu2019remote}. This work incorporated a recurrent neural network (RNN), specifically long short term memory (LSTM) over the temporal domain. For tasks such as heart rate detection and pulse detection, modest gains were observed over DeepPhys. The addition of the LSTM also allowed the network to be trained on the task of atrial fibrillation detection.

\section{Approach}
Using raw thermal videos, our emission representation model generates the input for the spatial-temporal network. This network, along with the detection network predicts the ISTI signal from the raw input thermal videos. Our proposed model architecture is shown in Figure ~\ref{fig:model_architecture}.

\begin{figure}[t]
\begin{center}
\includegraphics[width=\linewidth]{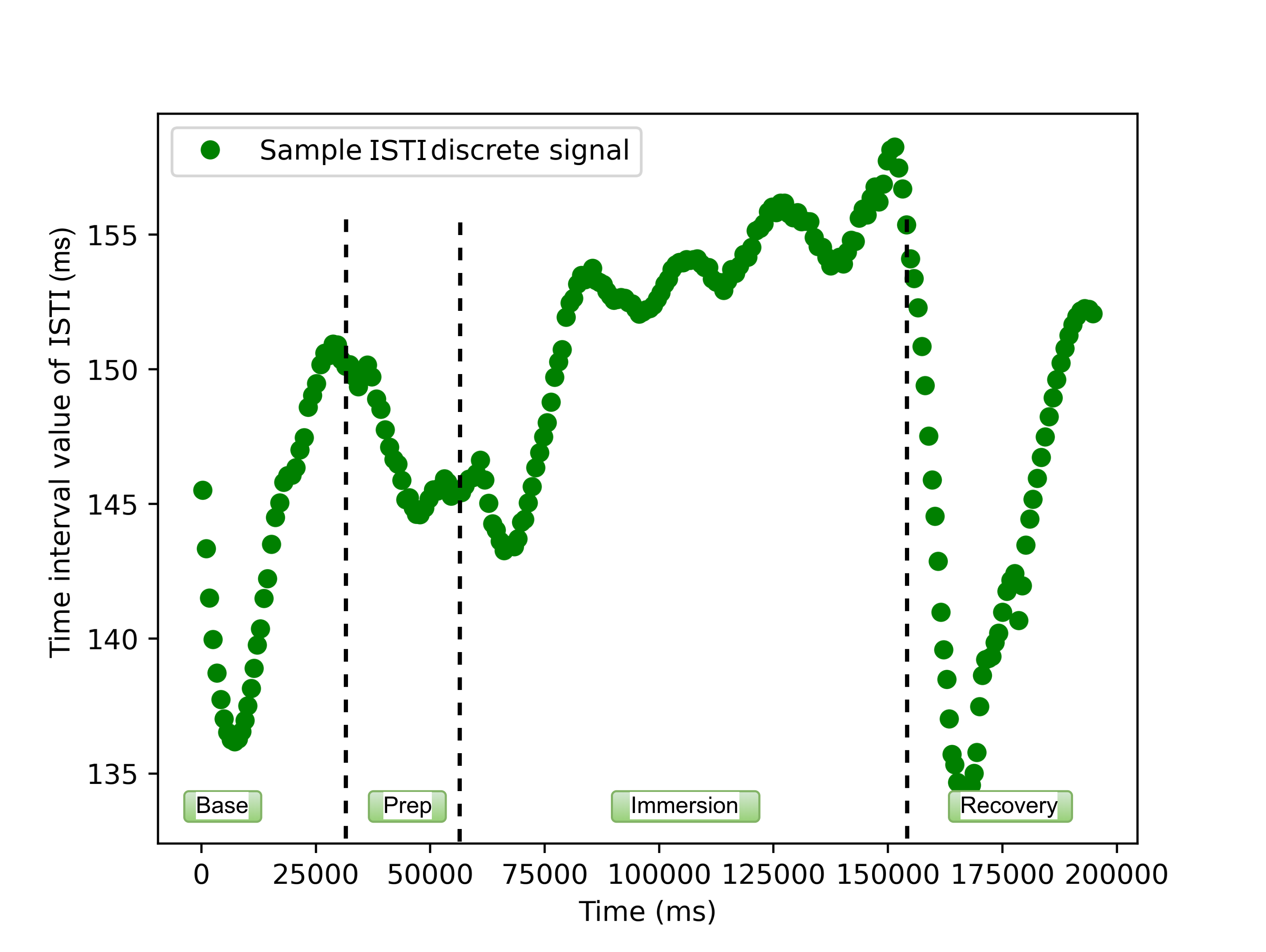}
\end{center}
   \caption{ Discrete ISTI values are plotted against the peak positions of the ECG signal for a single participant. The "Base", "Prep", "Immersion" and "Recovery" labels refer to different phases of our stress induction protocol, whereby participants immerse their feet in either ice-water ("stress" condition) or lukewarm water ("no-stress" condition). The data shown were randomly selected from the "no-stress" condition.  See section ~\ref{sec:dataset} for a detailed description of the protocol.
   }
\label{fig:pep}
\end{figure} 

\subsection{Generating ISTI signal}
Electrocardiography (ECG) and Impedance cardiography's (ICG or Z) derivative ($\partial Z / \partial t$) act as the gold-standard physiological signals. ISTI is defined as the interval from the onset of left ventricular depolarization, reflected by the Q-wave onset in the ECG, to the peak blood flow volume through aortic valve, reflected by the Z-point in the $\partial Z / \partial t$ signal. This time interval is computed from each peak of ECG and corresponding $\partial Z / \partial t$ peak. The discrete time interval value of ISTI is plotted at corresponding ECG peak positions as shown in Figure~\ref{fig:pep} and then interpolated with cubic interpolation to form a continuous signal. In Figure~\ref{fig:pep}, the x-axis represents time (ms) while y-axis values represent the ISTI value (ms) for a particular ECG peak at that time of the video. The interpolated continuous ISTI signal is used as the ground truth for ISTI prediction. 
%This interpolation is justified as the underlying stress controlling ISTI is slowly changing relative to heart rate.

\subsection{Emission Representation Model}
According to~\cite{wang2016algorithmic}, RGB video based physiological signal measurement involves modeling the reflection of external light by skin tissue and blood vessels underneath. However, in the case of thermal videos, the radiation received by the camera involves direct emissions from skin tissue and blood vessels, absorption of radiation from surrounding objects, and absorption of radiation by atmosphere~\cite{sanchez2009novel, ammer2004temperature}. Here, we build our learning model based on Shafer's dichromatic reflection model (DRM)~\cite{wang2016algorithmic} as it provides a basic idea to structure our problem of modeling emissions and absorption.
We can define the radiation received by the camera at each pixel location $(x, y)$ in the image as a function of time:

\begin{equation}
    \textit{\textbf{W}}^{x,y}(t) = E_{ems}^{x,y}(t) + E_{abs}^{x,y}(t) + E_{atm}^{x,y}(t)
    \label{eq:1}
\end{equation}
where $\textit{\textbf{W}}(t)$ is an energy vector (we drop the $(x,y)$ pixel location index in the following for simplicity.) $E_{ems}(t)$ is the total emissions from skin tissue and blood vessels; $E_{abs}(t)$ is absorption of radiations by skin tissue and blood vessels; $E_{atm}(t)$ is the absorption of radiation by atmosphere. In current experimental setup the person is in a closed environment and 3ft from thermal camera, therefore the atmospheric absorption is negligible. According to~\cite{hardy1934radiation}, human skin behaves as a black-body radiator, therefore the reflections are close to zero and emission is almost equivalent to absorption. 

This implies that the only variation in energy comes from the head motion and from blood flow underneath skin. If we decompose the $E_{ems}(t)$ and $E_{abs}(t)$ into stationary and time-dependent components:
\begin{equation}
    E_{ems}(t) = E_{o}\:.\:(\epsilon_s + \epsilon_b\:.\:f_1(m(t),\: p(t))
    \label{eq:2}
\end{equation}
where $E_o$ is the energy emitted by a black body at constant temperature, it is modulated by two components: $\epsilon_s$, is the emissivity of skin and $\epsilon_b$, is the emissivity of blood. $f_1(m(t),\: p(t))$ represents the variations observed by thermal camera; ~\cite{chen2018deepphys, wang2016algorithmic} $m(t)$ denotes all non-physiological variations like head rotations and facial expressions; $p(t)$ is the blood volume pulse (BVP). In a perfect black body, emissivity is equal to absorbtivity, therefore the absorbed energy is:
\begin{equation}
    E_{abs}(t) = E_{ab}(t)\:.\:(\epsilon_s + \epsilon_b\:.\:p(t))
    \label{eq:3}
\end{equation}
where $E_{ab}$ is the energy absorbed that changes with surrounding objects and their positions with respect to skin tissue.
\begin{equation}
    E_{ab}(t) = E_{o}\:.\:(1+ f_2(m(t),\:p(t)))
    \label{eq:4}
\end{equation}
where $f_2(m(t),\:p(t))$ represents the variation observed by the skin tissue. After substituting (\ref{eq:4}), (\ref{eq:3}), (\ref{eq:2}) in equation (\ref{eq:1}) and fusing constants; then neglecting the product of $f_1$ and $f_2$ as it is generally complex non-linear functions. Neglecting product of varying terms, we get an approximate $\textit{\textbf{W}}(t)$ as :
\begin{equation}
\begin{split}
    \textit{\textbf{W}}(t) \approx K + E_{o}\:.\:\epsilon_b\:.\:(p(t) + f_1(m(t),\: p(t))) \\ + E_{o}\:.\:\epsilon_s\:.\:f_2(m(t)\:, p(t))
\end{split}
\end{equation}
where K is 2$E_{o}\:.\:\epsilon_s$. We can get rid of this constant by taking first order derivative in the temporal domain. 
\begin{equation}
\begin{split}
    \textit{\textbf{W}}'(t) = p'(t)\:.\:E_o\:.\:(\epsilon_b + \epsilon_b\:.\:\frac{\partial f_1}{\partial p} + \epsilon_s\:.\:\frac{\partial f_2}{\partial p}) \\
    + m'(t)\:.\:E_o\:.\:(\epsilon_b\:.\:\frac{\partial f_1}{\partial m} + \epsilon_s\:.\:\frac{\partial f_2}{\partial m})
\end{split}
\end{equation}
This representation encompasses all the factors contributing to variations in radiation due to blood and face motion captured by the camera. Thus, we can suppress all possible non-necessary elements from data recorded by the camera.
% We should think of a better justification for this, I will read on an let you know if I think of anything
We use $\log$ non-linearity on each pixel to suppress any outlier in each image and separate the $E_{o}$, as its spatial distribution does not contribute to the physiological signal. The non-linearity looks as follows.
\begin{equation}
    \textit{\textbf{X}}(t) = sign(\textit{\textbf{W}}'(t))\:.\:\log(1 + \mod \textit{\textbf{W}}'(t))
\label{eq:7}
\end{equation}
To remove high frequency components, we do a Gaussian filtering with $\sigma = 3$ in the spatial domains, and $\sigma = 4$ in the temporal domain. This filtered $\textit{\textbf{X}}(t)$ is the input to our deep learning model. 

\subsection{Deep Learning Model}
\vspace{0.2cm}
\textbf{Spatial-Temporal Network:}
Spatial-Temporal networks are highly successful in action detection and recognition tasks ~\cite{ulutan2020actor, ulutan2020vsgnet, wang2014video}. More recently, such networks have been used to process multispectral signals~\cite{zhang2016feature,teffahi2019novel,kumar2020deep}
The input to our spatial-temporal network is the stacked features from the emission representation model, which are then fed to a backbone network (e.g. resnet-50~\cite{he2016deep}). This backbone network serves as a feature extractor. We mainly tested with object detection networks without the classification blocks as backbone networks.

Weights of these backbone networks are initialized with ImageNet pretrained values so that they can converge quickly on thermal videos. Global average pooling operation follows by the backbone block.

\begin{figure}[t]
\begin{center}
\includegraphics[width=\linewidth]{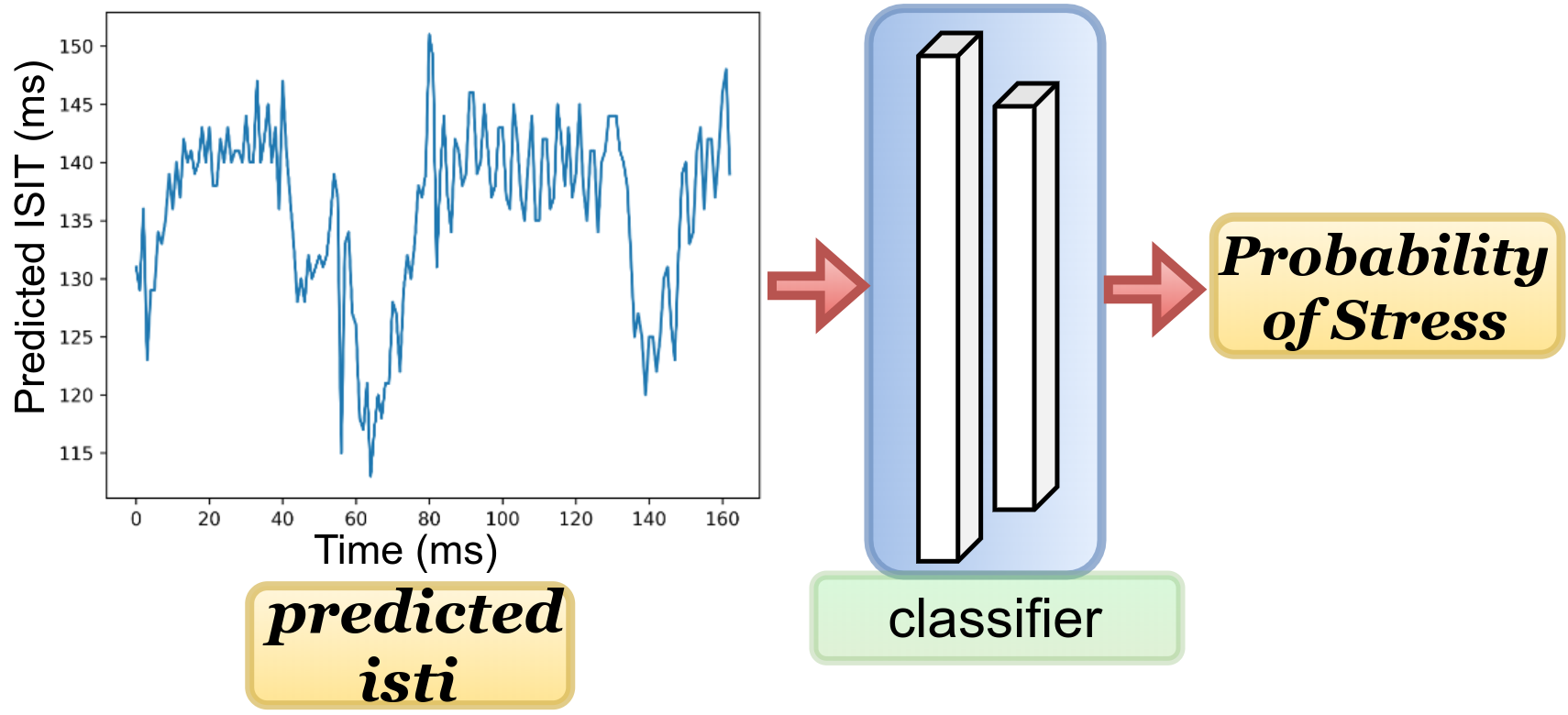}
\end{center}
   \caption{ Stress detection network. Estimated ISTI signal is directly fed into the classifier network to predict the probability that the subject is under stress.
   }
\label{fig:stress_detector}
\end{figure} 

\begin{equation}
    \textbf{f}_{0} = GAP(B\:(\textit{\textbf{X}}(t))
\end{equation}
where $B(\:.\:)$ stands for the backbone network, $GAP$ is global average pooling operation, $\textit{\textbf{X}}(t)$ is from equation[~\ref{eq:7}] (all processed frames stacked horizontally) and $\textbf{f}_o$ is the output feature vector.

The backbone network is followed by long short term memory (LSTM)~\cite{sundermeyer2012lstm, greff2016lstm} network, which captures the temporal contextual connection information from the extracted spatial features. LSTM~\cite{schuster1997bidirectional} units include a 'memory cell' that capture long range temporal context. A set of gates is used to control the flow of information which in turn helps the LSTM network learn temporal relations among the input features. The extracted feature vector from the backbone network is fed to the LSTM network.
%when information enters the memory, when it's output, and when it's forgotten. 
\begin{equation}
    \textbf{l}_o = L_{STM}\:(\textbf{f}_{0})
\end{equation}
where $\textbf{l}_o$ is the feature output from LSTM network, $L_{STM}(\:.\:)$ stands for LSTM network and $f_{0}$ is the extracted feature vector from the backbone network.

\vspace{0.3cm}

\textbf{Detection Network:}
Instead of directly predicting the continuous value of the ISTI signal from the output of LSTM network, we have divided the whole range [0,1] of ISTI values in $n$ number of bins following~\cite{ruiz2018fine}. To obtain the exact value of the ISTI signal ($\widehat{\textbf{ISTI}}$) from each frame the expectation of the probability is taken for over all bins ($\widehat{\textbf{isti}}_{bins}$),

%Experimenting with size, 6 layers of LSTM network used for capturing better temporal context information. The output feature vector from the resnet module is average-pooled and reshaped such that each single input sequence to LSTM module contains one second of temporal information(equation [~\ref{eq:9}]).This helps captures the sparsity of the PEP signal. PEP is a very sparse signal, that means PEP signal is available only when the ECG and ICG peaks occurs(i.e. once in one second). The output of the LSTM is further fed to two fully connected layers.
 
\begin{equation}
\begin{split}
    \widehat{\textbf{isti}}_{bins} = D \:(\textbf{l}_o);\:\:
    \widehat{\textbf{ISTI}} = \E\: (\widehat{\textbf{isti}}_{bins})
\end{split}
\label{pep_predictor}
\end{equation}
where \textit{D} stands for detection network which consists of fully connected layers, $\widehat{\textbf{isti}}_{bins} $ is the probability of each bin, $\E$ is the Expected value, $\widehat{\textbf{ISTI}}$ is the predicted ISTI value of each frame. This two stage approach makes our network more robust. 

%The final fully connected layers outputs multiple probability values per input frame. Two types of loss are computed here are explained in next section.
%\newline
The predicted $\widehat{\textbf{ISTI}}$ signal is fed to stress detection network which consists of fully connected layers, see Figure~\ref{fig:stress_detector}. The output of this network is probability of stress for the subject whose ISTI signal is estimated by our spatial-temporal network. %The stress detection network is shown in Figure~\ref{fig:stress_detector}. 
% \begin{equation}
%     \widehat{\textbf{cls}} = S_D\:(\widehat{\textbf{ISTI}})
% \label{stress_detector}
% \end{equation}
%where $S_D$ is the stress-detection network. Each detection network in equations ~\ref{pep_predictor} and ~\ref{stress_detector} includes non-linearity.

\subsection{Multi Loss Approach}

Previous works which predicted heart rate, breathing rate, or blood volume pulse mostly use mean squared error (MSE) loss.
% PEP signal or ECG and ICG? I think you can argue that we do not care about linear shift or scaling (i.e. what pearson normalizes out)
%Since the ISTI signal is non periodic.
%There are high chances of outliers and MSE is very sensitive to outliers values as pointed out by ~\cite{ruiz2018fine}.
Another approach bins the regression output, and modifies the network output layer to be a multi-class classification. This method provides more stability to outliers than MSE, but its accuracy is limited by the number of bins.
%MSE is more stable when combined with binned classification loss.
So for the ISTI signal prediction model, we use the multi loss approach used by~\cite{ruiz2018fine}. This type of loss is a combination of two components: a binned ISTI classification and an ISTI regression loss.
%We use the multi loss approach used by~\cite{ruiz2018fine}, which sets loss as a weight mean of an MSE loss and a binned regression loss.
%The intuition behind this type of loss is that by doing bin classification with a stable softmax and cross-entropy, it forces the network to learn to predict the neighbouring ISTI values. To refine the predictions, expectations of each output ISTI is computed for the binned ISTI. Next a regression loss is computed i.e. mean squared error, to improve the refined predictions.
%Better predictions of ISTI values are achieved by adding weight of 1 to classification loss and 0.7 to regression loss.
\begin{equation}
\begin{split}
    \textbf{L}\:(\Theta) = BCE\: (\widehat{\textbf{isti}}_{bins},\: \textbf{isti}_{bins})\:\: + \\\alpha\:.\:MSE(\widehat{\textbf{ISTI}},\: \textbf{ISTI})
\end{split}
\end{equation}

For the stress detection network only binary cross entropy (BCE) is used as loss function.

\begin{figure}[t]
\begin{center}
\includegraphics[width=\linewidth]{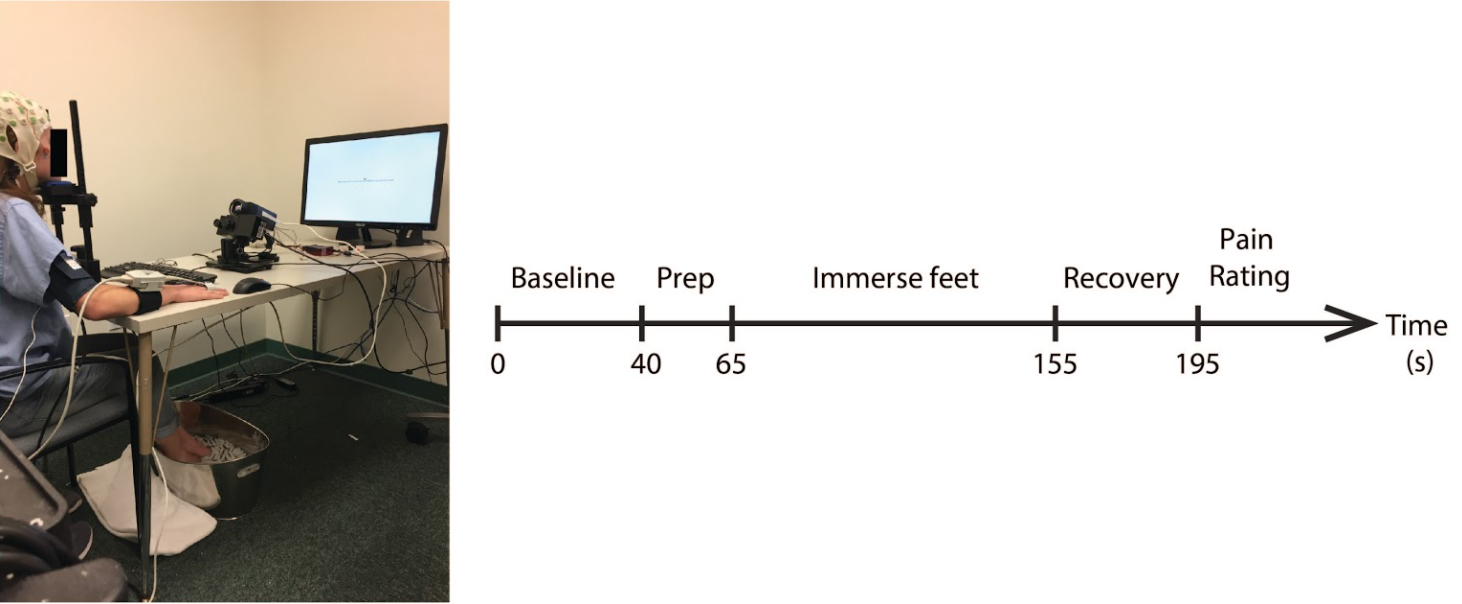}
\end{center}
   \caption{ CPT/WPT Setup and Protocol.  An example of a fully instrumented participant is shown.  Participants followed instructions for the protocol presented on a computer monitor.  After the baseline period the participant is instructed to position both feet on the edge of the bucket and prepare for immersion (prep).  They then immerse the feet for 90s, then withdraw the feet and rest them on a towel for a 40 s recovery period.              %\vspace{-0.5cm}
   }
\label{fig:dataset}
\end{figure} 

\section{Experiments}
%\vspace{-0.20cm}
\subsection{Dataset}
\label{sec:dataset}
\vspace{-0.20cm}

42 healthy adults (22 males, mean age 20.35 years) were recruited as part of the Biomarkers of Stress States (BOSS) study run at UC Santa Barbara, designed to investigate how different types of stress impact human brain, physiology and behavior. Participants were considered ineligible if any of the following criteria applied: heart condition or joint issues, recent surgeries that would inhibit movement, BMI $>$ 30, currently taking blood pressure medication or any psychostimulants or antidepressants. Informed consent was provided at the beginning of each session, and all procedures were approved by Western IRB and The U.S. Army Human Research Protection Office, and conformed to UC Santa Barbara Human Subjects Committee policies. 

Participants attended the lab for five sessions on five separate days as part of the BOSS protocol. For collection of impedance cardiography (ICG), 8 electrodes were placed on the torso and neck, two on each side of the neck and two on each side of the torso. For electrocardiogram (ECG), 2 electrodes were placed on the chest, one under the right collarbone.
For videos, thermal camera (Model A655sc, Flir Systems, Wilsonville, OR, USA),was positioned $\sim$65 cm from the participant's face and set to record at 640 $\times$ 240 pixels and 15 Hz frame rate. A large metal bucket was then positioned in front of the participant's feet.  In the Cold Pressor Test (CPT) session, the bucket was filled with ice water ($\sim$ 0.5 $^\circ$ C), whereas the in the control session (Warm Pressor Test; WPT), the bucket was filled with lukewarm water ($\sim$ 34 $^\circ$ C). In each session, participants were required to immerse their feet in the water five times for 90 s, following the test protocol outlined in Figure 5. The CPT is popular method for inducing acute stress in humans in the laboratory. It causes pain and a multiplex of physiological responses e.g. elevated heart rate and blood pressure and increased circulating levels of epinephrine and norepinephrine~\cite{zhang2018emotional, bachmann2018validation}. The WPT was devised as an "active" control task, designed such that participants engaged in exactly the same protocol as with the CPT, but without the discomfort of cold-water immersion. This ensured that any psychological or physiological effects induced by engaging in the protocol and immersing the feet in water, were controlled for. Each of the five CPT/WPT immersions were separated by $\sim$ 25 minutes. Between immersions, participants completed tests designed to measure performance across a range of cognitive domains (these data are not reported in this paper). Session order was counterbalanced between participants.  

Nine participants' data were excluded due to technical failures (the thermal imaging camera failed to record one or more sessions).  Thirty-three participants' data were used for modeling.  This sample is similar in size to existing public data sets of a similar nature ~\cite{soleymani2011multimodal, jaiswal2020muse}.

\subsection{Evaluation Metrics}
Performance metrics for evaluating ISTI prediction are Mean Squared Error (MSE) and Pearson's correlation coefficient (R). For stress detection, average precision (AP) is used as the validation metric. 

Mean Squared Error is a  model evaluation metric used for regression tasks. The main reason for using MSE as evaluation metric is that the precise value of predicted ISTI signal is important.

Pearson Correlation coefficients are used in statistics to measure how strong a relationship is between two signals. It is defined as covariance of the two signals divided by the product of their standard deviations. Pearson correlation is also used here as an extra validator on the predicted ISTI signal, signifying that the shape of predicted curve also corresponds well with the ground-truth.
\begin{equation*}
    \rho_{X,Y} = \frac{cov(\:X, \:Y)}{\sigma_X \:.\: \sigma_Y}
\end{equation*}

Average Precision (AP) is the most commonly used evaluation metric for object detection tasks~\cite{everingham2010pascal}. It estimates the area under the curve of precision and recall plot. Precision measures how many predictions are correct. Recall calculates the correctly predicted portion of the ground truth values.     

\subsection{Implementation Details}
In experiments, the effectiveness of the spatio-temporal network is evaluated. The dataset is split as follows: 80\% training, 10\% validation and 10\% testing set.
The input video frames are cropped to 360 $\times$ 240 to remove the lateral blank areas before being fed to our emission representation model. 

For backbone model, experiments with different architectures of resnet were performed, those are resnet18, resnet34, resnet50, resnet101. In the final model resnet50 is used as feature extractor. The output of resnet50 is average-pooled instead of max-pooling operation. The reason for that is removing a less important feature from important feature (max-pool operation) can reduce the signal-to-noise ratio in physiological measurement, so average pooling is used to keep even the less important feature vector information. Before feeding to temporal network, the average-pooled feature vector is reshaped so that each input sequence to LSTM consists of 1 second of time information. The reasoning was that since the peaks of ECG and $\partial Z / \partial t$ signal occurs almost once per second, the LSTM network will better captures the relation between adjacent peaks.
For the temporal network, we experimented with a number of LSTM layers (2-8), 6 LSTM layers are best suited for capturing the temporal contextual information. Hidden unit size is kept at half the feature vector length from the spatial network (resnet50), $hidden\_unit\_size$ is 256 $\times$ $frame\_rate$, ensuring that the number of memory cells is sufficient enough to transfer information from previous LSTM cell to next. 
The number of fully connected layers following LSTM is two, with ReLU added as non-linearity. The output of the final fully connected layer is 33 bins output. 33 bins is an empirical value.

The emission representation model works online in the pipeline and is loaded on the same machine on which deep learning model is trained. Each video is approximately of size (frames $\times$ H $\times$ W) 2500 $\times$ 640 $\times$ 240 with 16bit depth information per pixel. Due to memory constraints on the GPU, $batch\_size$ is kept at 500 frames. The learning rate for resnet50 is started at 0.001, for LSTM and FC layers at 0.01, which reduces after every 10 epochs by a factor of 0.1. Stochastic Gradient Descent is used as optimizer for the network.

\begin{table}[t]
\begin{tabular}{|l|r|r|}
\hline
Name of the Method                          & \multicolumn{1}{l|}{PC Coefficient} & \multicolumn{1}{l|}{MSE} \\ \hline
Baseline                                    & 0.170                                              & 103.829                 \\ \hline
DeepPhys ~\cite{chen2018deepphys}                                    & 0.575                                              & 47.530                  \\ \hline
I3D ~\cite{carreira2017quo} + Detection Network   & 0.84                                             & 5.227                \\ \hline
\textbf{StressNet}                                   & \textbf{0.843}                                     & \textbf{5.845}          \\ \hline
\end{tabular}
\vspace{0.20cm}
\caption{StressNet's performance in predicting ISTI signal. The performance is measured on Pearson-Correlation Coefficient(PC Coefficient) and mean square error. Our model clearly outperforms the existing methods by a good margin.}
\label{tab:ISTI_per}
\end{table}

\section{Results}
The proposed method is evaluated in two main criteria. First we evaluated the quality of our predicted ISTI signal, then we tested the effectiveness of the predicted ISTI signal in detecting stress. 
%The proposed method is evaluated on two main aspects, first is the evaluation of predicted initial systolic time interval (ISTI) signal and second is the validation that ISIT is more effective for stress detection. 

\begin{figure}[t]
\begin{center}
\includegraphics[width=\linewidth]{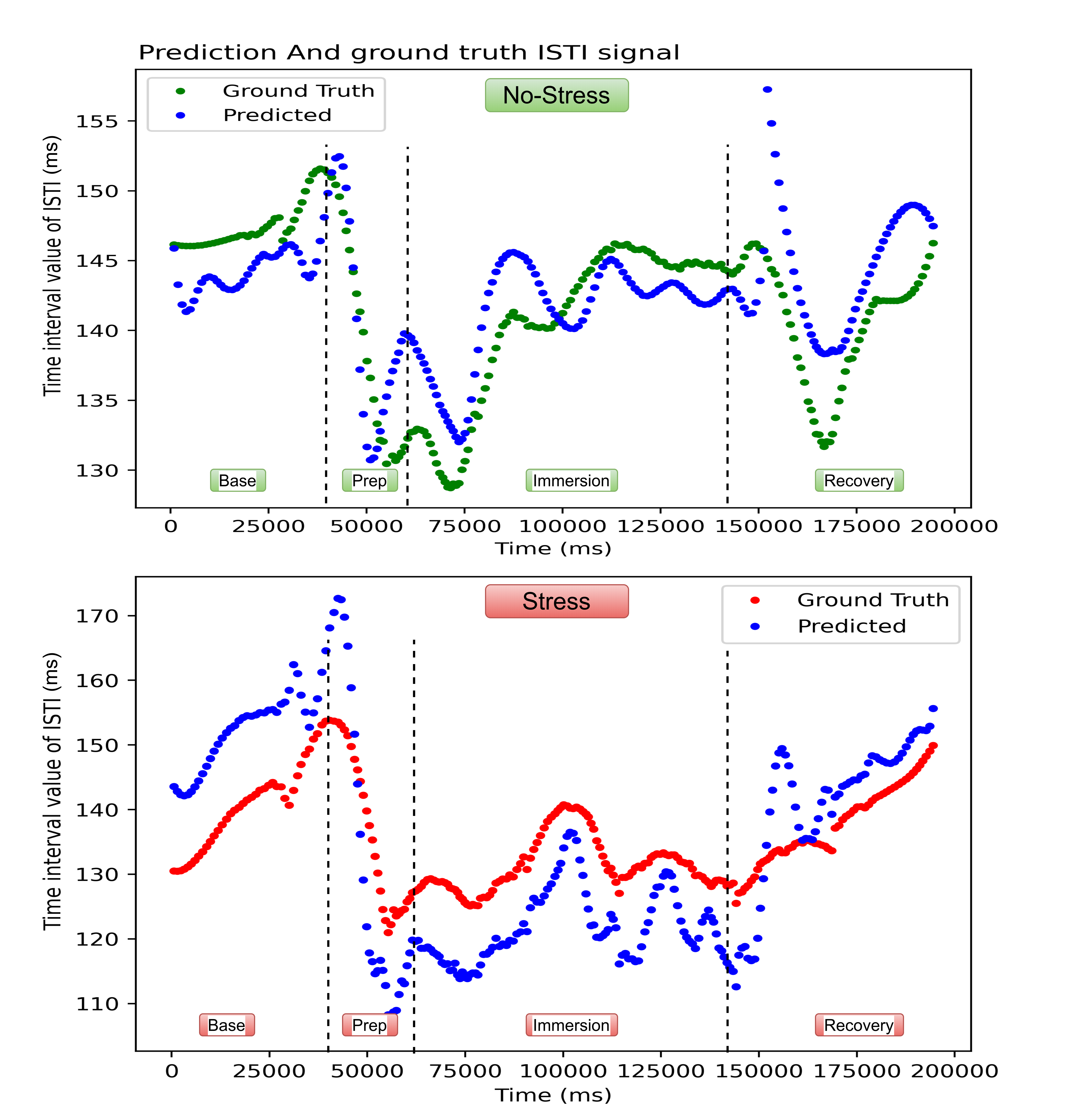}
\end{center}
   \caption{Quality of our predicted ISTI signal in stress and no-stress conditions. Data shown are examples from a single participant's data (selected at random).  The "Base", "Prep", "Immersion" and "Recovery" labels refer to the different phases of the CPT/WPT procedure.%\vspace{-0.5cm}
   }
\label{fig:isti_pred}
\end{figure}

%\vspace{0.3cm}

\textbf{Predicting ISTI Signal:}
For the first part, as mentioned in the evaluation metrics section, the model performance is evaluated on Mean Squared Error (MSE) and Pearson Correlation coefficient (PC Coefficient). In Table~\ref{tab:ISTI_per} our model's performance can be seen compared to the other methods. Our model outperforms the other methods in both of the evaluation metrics with a good margin. As shown in Figure~\ref{fig:isti_pred}, our model agrees well with the ground truth signal in both stress and no-stress cases.
%For the first part, on training and testing with different backbone architectures, the emission representation model along with spatial-temporal network and detection network predicted the ISTI signal with a Mean Squared Error of ... and Pearson Correlation coefficient(R) of ... 

Since no work has been done on detecting the ISTI signal before, to validate our model we have implemented DeepPhys~\cite{chen2018deepphys}. As can be seen in Table~\ref{tab:ISTI_per} our implementation of DeepPhys model~\cite{chen2018deepphys} did not perform well in detecting the ISTI signal. This poor performance mostly stems from two main reasons. First, DeepPhys model is designed to predict periodic physiological signals and since ISTI is non-periodic in nature, loss in DeepPhys does not suit this particular task. Second,  the skin reflection model in~\cite{chen2018deepphys} does not expand properly for modeling the infrared radiation.
%Also, we have implemented I3D ~\cite{carreira2017quo} architecture, a 3D convolution based action recognition network, to compare with our network. We have replaced the classification branch with our detection network in I3D. It also performs poorly compared to our network. One reason may be I3D gets overwhelmed by the redundant information coming from consecutive frames as it directly aggregates feature information from the first layer of the network.    
%the infrared radiation well, considering the fact the human body act as a perfect black body~\cite{hardy1934radiation}, there are no reflections, just absorptions and emissions only. 
%We also compare our work with Yu et al. ~\cite{yu2019remote}. Yu et al. developed two spatial-temporal network: one is CNN+LSTM based model, the other one is similar I3D~\cite{carreira2017quo} network.   But our model leverages the deeper residual network with strong loss function. The loss function proposed by~\cite{yu2019remote} relies only on matching trend in the signal. This does not perform well when the target signal is non-periodic.  
For baseline methods, ECG signal is extracted from the face using simple statistical filtering methods. According to~\cite{engert2014exploring, genno1997using, veltman2005facial} temperature changes in the tip of the nose and forehead can index different stress states, so for our baseline approach we tracked these regions and then band-pass filtered to extract the signal. This signal is quite noisy which contributes to our baseline's poor performance.

\begin{figure}[t]
\begin{center}
\includegraphics[width=\linewidth]{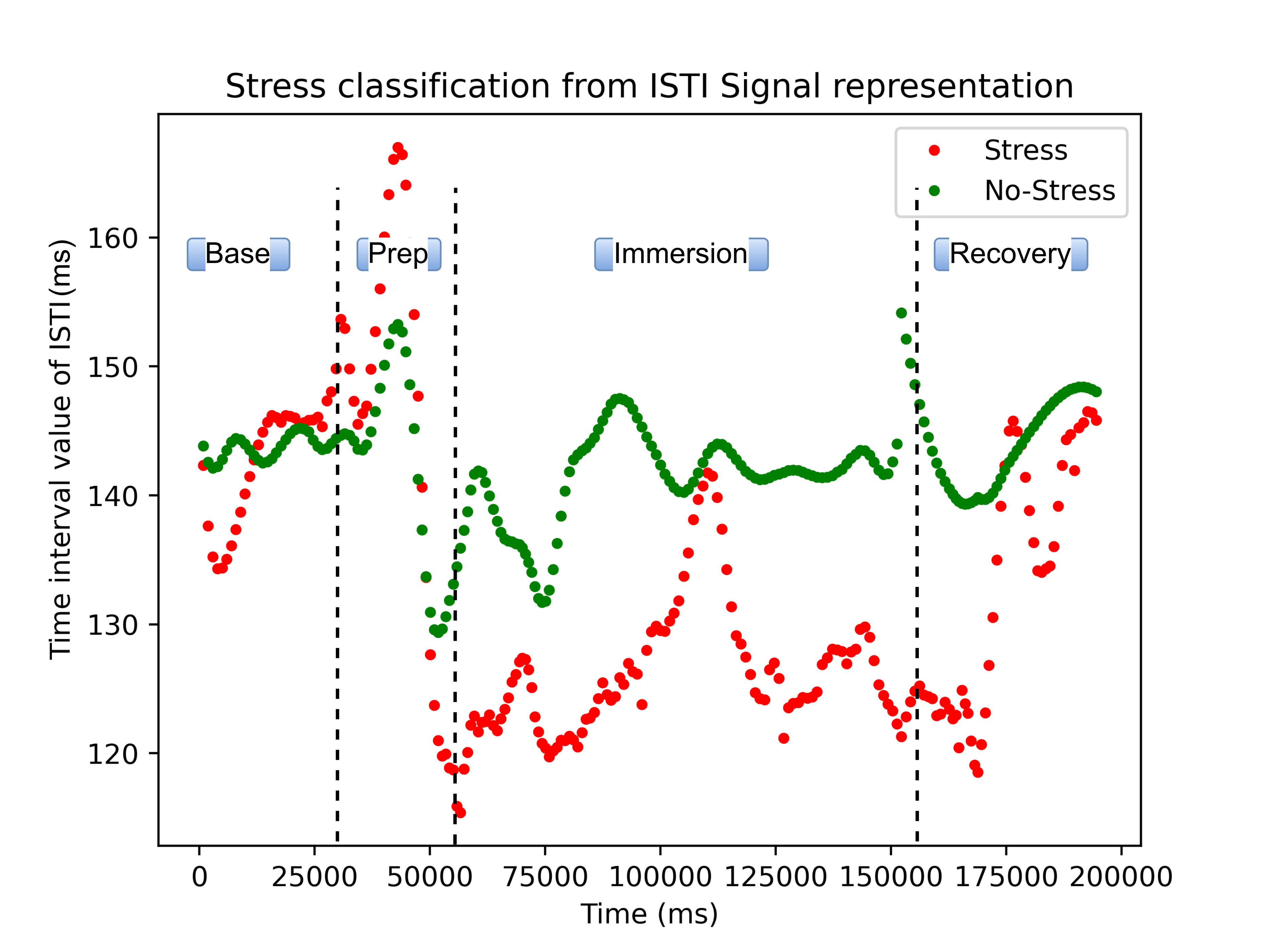}
\end{center}
   \caption{Importance of ISTI signal in detecting stress. Ground truth ISTI data from a single participant (randomly selected) are shown. Clearly, ISTI signal in the stress condition is different from the ISTI signal in no-stress condition.  The "Base", "Prep", "Immersion" and "Recovery" labels refer to the different phases of the CPT/WPT procedure. 
   %The big peak and dip in the ISTI signal actually matches with the time stress is induced on the subject.             %\vspace{-0.5cm}
   }
\label{fig:s/ns-cls}
\end{figure} 

\begin{table}[]
\begin{tabular}{|l|l|}
\hline
Input Signal                 & AP (Average Precision)               \\ \hline
Heart Rate (HR)             & \multicolumn{1}{r|}{0.753}       \\ \hline
Heart Rate Variability(HRV) & \multicolumn{1}{r|}{0.814}        \\ \hline
ISTI (Ground Truth Signal)  &  \multicolumn{1}{r|}{0.902}       \\ \hline
\textbf{ISTI (StressNet Predicted)}  & \multicolumn{1}{r|}{\textbf{0.842}} \\ \hline

\end{tabular}
\vspace{0.20cm}
\caption{StressNet can classify stress state with greater AP using contact-less ISTI estimates when compared to other commonly used contact-less signal estimates (HR and HRV).  }
\label{tab:stress_det}
\end{table}
%\vspace{0.3cm}
\textbf{Detecting Stress:}
For the  second part of stress detection, we evaluate whether the ISTI signal provides a robust index for stress detection. %our stress detection model achieved an Average precision of 0.84.

An example of the ISTI response to CPT/WPT in a single participant is shown in Figure ~\ref{fig:s/ns-cls}. Here, we observe a clear distinction in ISTI in anticipation of cold- vs. warm- water immersion (i.e. during the prep period) as well as during immersion and recovery.  

%As shown in Figure ~\ref{fig:s/ns-cls}, qualitatively ISTI signal is quite significant when detecting stress, both mental and physical. The green curve represents no-physical stress and the red curve represents physical stress induced. A distinction is clearly visible, important thing to notice is that, the red curve shows an initial fall in ISTI value, it shows the extra mental pressure because the person was made aware of that he/she has to put his/her feet in cold water at certain point in time.

To evaluate the predictive validity of the ISTI signal, we compare it to heart rate ~(HR) and heart rate variability ~(HRV) by entering these alternative signals into our model. We compare ISTI with HR and HRV because these measures are considered to be reliable indices of stress~\cite{spellenberg2020binary} and have been used in many stress classification studies~\cite{bousefsaf2013remote, pluntke2019evaluation}.
Here, we compute them from the ground truth ECG signal. HR is computed by counting number of beats in a sliding window approach with window size 15~(seconds) and stride 1~(seconds). For HRV, time between R peaks is recorded over a defined time interval~(15 seconds) and then HRV is computed according to the Root Mean Square of Successive Differences~(RMSSD) method~\cite{Bryn:2019}. 

In Table ~\ref{tab:stress_det} we can see how our predicted ISTI signal is better in detecting stress state than HR (12\% higher AP) and HRV (4 \% higher AP). Also, higher AP with the ground truth ISTI signal confirms that ISTI is the most reliable index of stress state in the context of our dataset.
%Also, We have used ground truth ISTI signal as input to our stress detection network. 
\begin{table}[]
\begin{tabular}{|l|l|l|}
\hline
Name of the Backbone & PC Coefficient              & MSE                         \\ \hline
vgg19 ~\cite{simonyan2014very}            &  \multicolumn{1}{r|}{0.605}                & \multicolumn{1}{r|}{33.164}                 \\ \hline
resnet18 ~\cite{he2016deep}             & \multicolumn{1}{r|}{0.749}                          & \multicolumn{1}{r|}{15.095}                          \\ \hline
resnet34 ~\cite{he2016deep}            & \multicolumn{1}{r|}{0.815} & \multicolumn{1}{r|}{6.223} \\ \hline
\textbf{resnet50} ~\cite{he2016deep}             & \multicolumn{1}{r|}{\textbf{0.843}}                          & \multicolumn{1}{r|}{\textbf{5.845}}                          \\ \hline
resnet101 ~\cite{he2016deep}            &  \multicolumn{1}{r|}{0.779}                & \multicolumn{1}{r|}{14.373}                 \\ \hline
%i3d                  & NN                          & NN                          \\ \hline
\end{tabular}
\vspace{0.30cm}
\caption{Comparison of different backbones' performance. In the task of estimating ISTI signal resnet50 is better than all other backbones. }
\label{tab:ISTI_per_back}
\end{table}
\subsection{Ablation Study}
%\vspace{0.3cm}
\textbf{Analysis of Emission Representation Model:} The overall architecture of StressNet consists of three main models: the emission representation model, the spatial-temporal model and the detection model. To evaluate how each model affects the overall performance, we evaluated the spatial-temporal model with and without the emission representation model. The fully pre-trained network was tested without the emission representation model and we observed a 1.119 increase in the mean squared error in predicting the ISTI signal. The best results for ISTI signal prediction as mentioned in table~\ref{tab:ISTI_per} are obtained using all three models mentioned above.
%\vspace{0.3cm}

\textbf{Analysis with Backbone CNNs:} The spatial-temporal model is evaluated with all the ResNet models~\cite{he2016deep}. %We also replaced the whole ResNet+LSTM with I3D model for evaluation of spatial-temporal model. 
We also tested with VGG19~\cite{simonyan2014very} as our backbone. The performance comparison is shown in table~\ref{tab:ISTI_per_back}.
%\vspace{0.3cm}

\textbf{Analysis with Breathing signal:} The breathing signal is captured by tracking the area under the nostrils for changes in temperature. The computed time series signal is passed through band-pass filter with low and high cutoff frequencies of 0.1 Hz and 0.85 Hz, respectively. This breathing signal is also used as an input to our stress detection model and the predictions from this model are multiplied with the predicted ISTI signal input. This process boosts the stress detection results by 0.1774 AP. This shows how ISTI can be combined with other physiological signals to detect stress.
%\vspace{0.3cm}

\begin{figure}[t]
\begin{center}
\includegraphics[width=\linewidth,height=3cm]{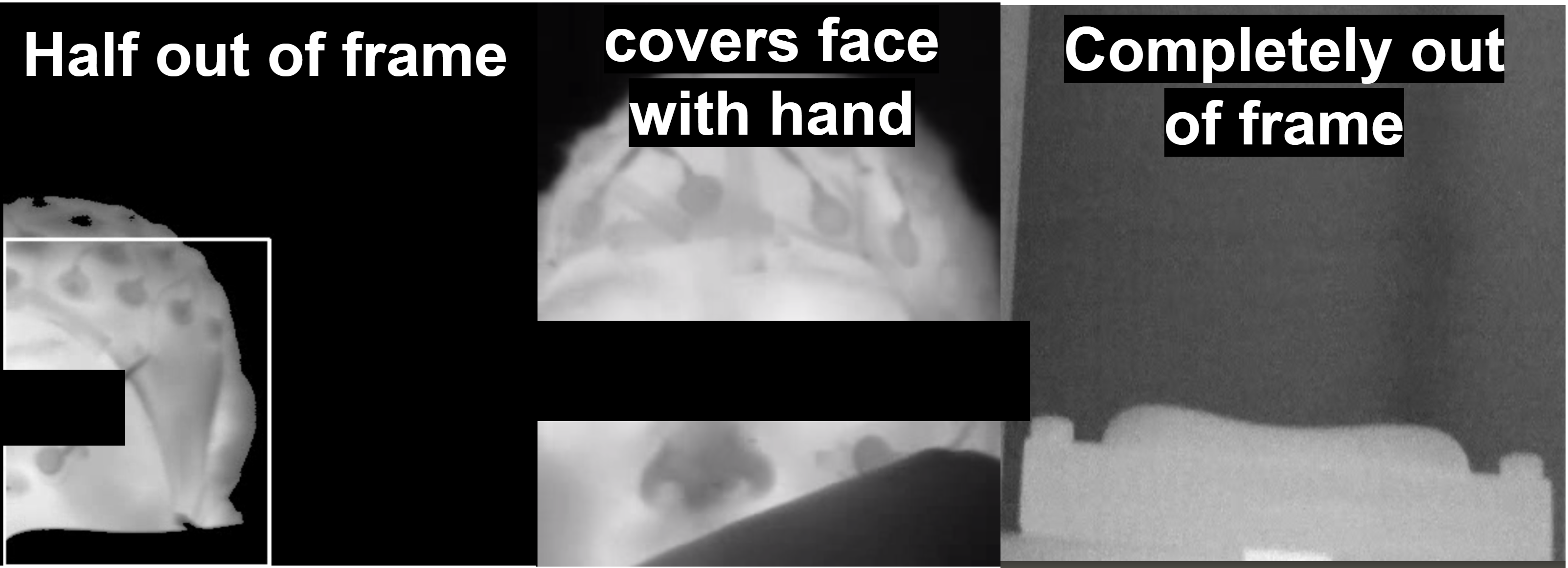}
\end{center}
   \caption{ Example StressNet failure cases. Network performance is impaired when the face is outside the video frame or obscured. 
   %The top part shows the noise in dz/dt signal and the bottom part shows the person out of the frame.
   }
\label{fig:failcase}
\end{figure}

\textbf{Limitations of the Model:} Despite being instructed to stay still, participants occasionally made large head movements and/or obscured their face with a hand (see Figure~\ref{fig:failcase}). There were also occasions where the ECG/ICG signal was noisy due to movement or bad electrode connections. In these instances the model fails to detect ISTI.
%\vspace{0.3cm}

\textbf{Different Spatial Temporal Network:} To validate the effectiveness of spatial temporal networks in detecting ISTI signal, we implemented I3D ~\cite{carreira2017quo} architecture, a 3D convolution based spatial-temporal network proposed for action recognition. We replaced the classification branch in I3D with our detection network. 
The performance is similar to StressNet's performance.
%The results are shown in Table ~\ref{tab:spa_net}. The slight better performance of I3D ~\cite{carreira2017quo} in terms of MSE error that the future research can focus on finding a better spatial-temporal network.
%It also performs poorly compared to our network. One reason may be I3D gets overwhelmed by the redundant information coming from consecutive frames as it directly aggregates feature information from the first layer of the network.     
\vspace{-0.2cm}
\section{Conclusion}

Here we present a novel method for the estimation of ISTI from thermal video and provide evidence suggesting ISTI is a better index for stress classification than HRV or HR. Overall, our method is more accurate than existing methods when performing binary stress classification on thermal video data. 

Our model achieved state-of-the-art performance, and performance could potentially be boosted even further by using different spatial-temporal models. The most successful backbone model used only spatial data from each frame independently, compared to the I3D network~\cite{carreira2017quo} that employed simultaneous processing of both spatial and temporal information. However, to test this we require larger dataset, that would allow for improved pre-trained initialization of the spatial-temporal backbones and better transfer learning  performance.

This work has several limitations. First, it is unclear whether StressNet's performance can generalize to the classification of different forms of stress e.g. social stress, physical and mental fatigue. Second, it is possible that exposure to lukewarm-water in the control condition may have induced eustress (beneficial stress), meaning that StressNet is actually classifying distress vs. eustress, not distress vs. neutral states, and this may impact performance. Third, the data used to test StressNet were collected under controlled laboratory conditions, so it is unclear how performance may be impacted in real world use case scenarios that may by subject to increased atmospheric noise and movement artifacts. Further testing with a diverse range of datasets collected under different stress conditions and scenarios is required to determine the efficacy and generalizability of StressNet in the real world.

\section{Acknowledgements}
This research is supported in part by NSF Award number 1664172 and by the SAGE Junior Fellows Program. The dataset was collected for the UC Santa Barbara Biomarkers of Stress States project, supported by the Institute for Collaborative Biotechnologies through contract W911NF-09-D-0001, and cooperative agreement W911NF-19-2-0026, both from the U.S. Army Research Office.

%In this work, we presented a novel method for estimation of ISTI from thermal video. We provide further evidence that ISTI may be a better metric for stress than HRV or AHR. As a whole, our method was more effective current methods at performing binary stress classification methods in binary stress classification from video inputs.

%Though our model achieved state-of-the-art performance, further gains could be potentially be achieved different spatial-temporal models. The most successful backbone model used only spatial data from each frame independently, compared to the I3D network which processed spatial and temporal information simultaneously. Larger video classification datasets in the near future may allow for better pretrained initializations of spatial-temporal backbones, and better transfer performance on our task.

%Our findings indicate that StressNet can successfully predict stress state in individuals that are either exposed to a painful stimulus or not. Further experiments will test whether StressNet can be generalized to classify other types of stress in humans (e.g. social stress, physical or mental fatigue). With our success on ISTI, we are also interested in expanding our subject data. This may include additional stresses and combinations of stresses, and more physiological measurements.

{\small
\bibliographystyle{ieee_fullname}
\bibliography{01StressNet}
}

\end{document}